\documentclass{article}

\pdfoutput=1

    \usepackage[final,nonatbib]{neurips_2022}

\usepackage[utf8]{inputenc} 
\usepackage[T1]{fontenc}    
\usepackage{hyperref}       
\usepackage{url}            
\usepackage{booktabs}       
\usepackage{amsfonts}       
\usepackage{nicefrac}       
\usepackage{microtype}      
\usepackage{xcolor}         

\usepackage{amsmath}
\usepackage{graphicx}
\usepackage{mathabx}
\usepackage{caption}
\usepackage{subcaption}

\title{Debiasing Meta-Gradient Reinforcement Learning by Learning the Outer Value Function}

\author{%
  Cl\'{e}ment Bonnet~\thanks{Correspondence to: <\texttt{c.bonnet@instadeep.com}>} \\
  InstaDeep \\
  \And
  Laurence Midgley \\
  InstaDeep \\
  \And
  Alexandre Laterre \\
  InstaDeep \\
}

\begin{document}

\maketitle

\begin{abstract}
Meta-gradient Reinforcement Learning (RL) allows agents to self-tune their hyper-parameters in an online fashion during training.
In this paper, we identify a bias in the meta-gradient of current meta-gradient RL approaches.
This bias comes from using the critic that is trained using the meta-learned discount factor for the advantage estimation in the outer objective which requires a different discount factor.
Because the meta-learned discount factor is typically lower than the one used in the outer objective, the resulting bias can cause the meta-gradient to favor myopic policies.
We propose a simple solution to this issue: we eliminate this bias by using an alternative, \emph{outer} value function in the estimation of the outer loss. 
To obtain this outer value function we add a second head to the critic network and train it alongside the classic critic, using the outer loss discount factor.
On an illustrative toy problem, we show that the bias can cause catastrophic failure of current meta-gradient RL approaches, and show that our proposed solution fixes it.
We then apply our method to a more complex environment and demonstrate that fixing the meta-gradient bias can significantly improve performance.
\end{abstract}

\section{Introduction}

Recently, reinforcement learning (RL) methods have embraced ideas from meta-learning either to train an agent to quickly adapt to new tasks~\cite{finn2017,kirsch2019} or to improve the learning process online within a single task~\cite{meta_rl_xu,zheng2018,veeriah2019,xu2020,zahavy2020}. 
RL algorithms are known to be highly sensitive to their hyperparameters such as the discount factor~\cite{amit2020discount}. 
Thus, self-tuning these hyper-parameters online can greatly improve performance and has yielded state-of-the-art performance for model-free RL~\cite{flennerhag2022bootstrapped}.
Meta-gradient RL is the backbone underlying self-tuning and relies on an \emph{inner} loss function used to update the agent's parameters, as well as an \emph{outer} loss to evaluate (or ground) the updated parameters in order to compute a meta-gradient. This meta-gradient is then used to update the meta-parameters (e.g. the discount factor).

In this paper, we focus on the outer loss used by current meta-RL algorithms and show that its estimation has a bias.
This bias comes from using the critic that is trained using the meta-learned discount factor for the advantage estimation in the outer loss.
Meta-gradients obtained using this biased estimate can lead to a failure of self-tuning, potentially resulting in a myopic policy.
We propose a computationally cheap approach that eliminates the meta-gradient bias and we observe that it solves catastrophic failure in a toy environment and improves performance on a more challenging deep RL task.

There have been several works focusing on bias and variance in meta-gradients. (Bonnet et al., 2021)~\cite{bonnet2021one} focuses on bias-variance trade-off within multi-step meta-gradients. (Vuorio et al., 2022)~\cite{vuorio2022} studies a different bias, namely the sampling bias, and (Liu et al., 2021)~\cite{liu2021} addresses a compositional bias. 
Although these aforementioned works are complementary to ours, the bias we highlight is different as it is related to the outer loss of meta-gradient algorithms. 


\section{Background}
\label{section:background}
\textbf{Reinforcement Learning}:
The goal of an RL agent is to learn the policy $\pi$ that maximizes the $\gamma$-discounted expected return:  $\mathbb{E}_{\pi | s_t = s} \left[\sum_{k=0}^{\infty} \gamma^k r_{t+k}\right]$, where $r_t$ and $s_t$ are the reward and state at timestep $t$.
Actor-critic algorithms~\cite{konda1999,mnih2016} learn a policy $\pi_{\theta_p}$ with parameters $\theta_p$ and a critic $V^{\pi,\gamma}_{\theta_c}$ with parameters $\theta_c$ that approximates the value function $V^{\pi,\gamma} = \mathbb{E}_{\pi | s_t = s} \left[\sum_{k=0}^{\infty} \gamma^k r_{t+k}\right]$.
The critic is used to inform the direction of the policy update. 
More specifically, the policy is trained using the policy gradient ~\cite{Williams1992,gae_paper}, $\hat{g}_\textrm{pg}$, estimated by Monte Carlo  using
\begin{equation}
\label{eq:policy_gradient_estimate}
    \hat{g}_\textrm{pg}(\tau) = \left(r_t + \gamma V^{\pi,\gamma}_{\theta_c}(s_{t+1})\right) \nabla_{\theta_p} \log \pi_{\theta_p}(a_t|s_t)\,
\end{equation}
where $\tau = \{s_t,a_t, r_{t}, s_{t+1}\}$ are transitions.
One could use the advantage~\cite{gae_paper} instead of the action value function in the estimate to reduce variance~\cite{Greensmith2001,mohamed2019}, but we stick to the action value function for simplicity.
The full actor-critic update combines the policy gradient with entropy regularization~\cite{schulman2017,mnih2016,sac,gruslys2017} and a critic update using the mean squared error of the TD error~\cite{sutton2018}.
We provide further discussion and definitions in Appendix~\ref{appendix:background}.

\begin{figure}
     \centering
     \begin{subfigure}[b]{0.42\textwidth}
         \centering
         \includegraphics[width=\textwidth]{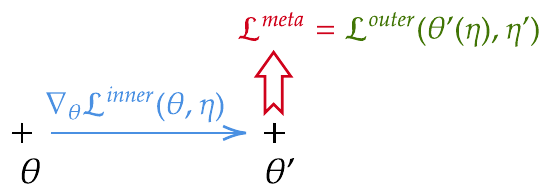}
         \caption{Meta-gradients (MG)}
         \label{fig:meta_loss_mg}
     \end{subfigure}
     \hfill
     \begin{subfigure}[b]{0.53\textwidth}
         \centering
         \includegraphics[width=\textwidth]{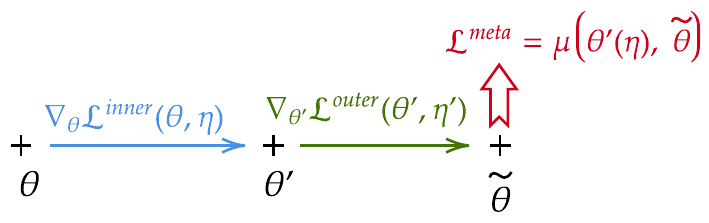}
         \caption{Bootstrapped meta-gradients (BMG)}
         \label{fig:meta_loss_bmg}
     \end{subfigure}
        \caption{The two paradigms of meta-gradients for self-tuning. First, parameters $\theta$ are updated using an inner loss and meta-parameters $\eta$. Then, a meta-loss is computed from the updated $\theta'$. Meta-gradients (MG) and bootstrapped meta-gradients (BMG) differ in their choice of meta-loss.}
        \label{fig:meta_loss}
\end{figure}

\textbf{Self-Tuning RL}:
We refer to the aforementioned actor-critic update as the \emph{inner update} with loss 
$\mathcal{L}^\textrm{inner}(\theta, \eta)$ where $\eta$ are the hyper-parameters.
Self-tuning meta-RL purposely exposes $\eta$ as trainable meta-parameters and adapts them online during training by computing meta-gradients.
This makes use of the fact that the inner update is differentiable with respect to $\eta$.
The meta-gradient $\nabla_\eta\mathcal{L}^\textrm{meta}$ is defined as the gradient of a meta-loss with respect to the meta-parameters. 
It is computed in two phases (see figure~\ref{fig:meta_loss}): \textbf{(1)} The agent takes one (or multiple) parameter update(s) by running the actor-critic algorithm with meta-parameters $\eta$ to obtain updated parameters $\theta'(\eta)$. \textbf{(2)} The meta-loss is computed from the updated parameters $\theta'(\eta)$ to derive the meta-gradient $\nabla_\eta \mathcal{L}^\textrm{meta}(\theta'(\eta))$.
One can distinguish two kinds of meta-loss (see figure~\ref{fig:meta_loss}). Firstly, \emph{meta-gradients} (MG)~\cite{meta_rl_xu,xu2020,zahavy2020,veeriah2019} compute the gradient of an outer loss $\mathcal{L}^\textrm{outer}$ applied to updated parameters $\theta'(\eta)$. Secondly, \emph{bootstrapped meta-gradients} (BMG)~\cite{luketina2022,flennerhag2022bootstrapped} further update the parameters by taking one (or several) gradient step(s) to obtain a target $\tilde{\theta}$ from which to compute a matching loss between $\theta'(\eta)$ and $\tilde{\theta}$. The target updates proposed by \cite{flennerhag2022bootstrapped} can be done with respect to the inner loss $\mathcal{L}^\textrm{inner}$ for all but the last step, which is computed with an outer loss $\mathcal{L}^\textrm{outer}$.
Both methods of computing meta-gradients necessitate an outer loss, $\mathcal{L}^\textrm{outer}$, that is defined to be an actor-critic loss similar to $\mathcal{L}^\textrm{inner}$ but uses different hyper-parameters $\eta'$ than the inner loss. In particular, the policy gradient term of the outer loss uses a different discount factor $\gamma'$, leading to the following policy gradient estimate $\hat{g}_\textrm{pg}'$.

\begin{equation}
\label{eq:outer_policy_gradient_estimate}
    \hat{g}_\textrm{pg}'(\tau) = \left(r + \gamma' V^{\pi,\gamma'}_{\theta'}(s')\right) \nabla_{\theta'} \log \pi_{\theta'}(a|s)
\end{equation}

\section{Solution to the meta-gradient bias}

In this section, we highlight a bias in the meta-gradient estimation of current meta-RL methods and propose a simple and computationally-efficient way to solve it.

\textbf{Bias in the meta-gradient}:
Both MG and BMG methods rely on the estimation of the outer loss to compute the meta-gradient (see figure~\ref{fig:meta_loss}).
As seen in section~\ref{section:background}, the policy gradient terms from the inner and outer loss differ in terms of what hyper-parameters they use. 
While the inner loss necessitates estimating the value function $V^{\pi,\gamma}$, the outer loss requires $V^{\pi,\gamma'}$.
Since the critic is trained to estimate the value function $V^{\pi,\gamma}$, a bias consequently arises when using the critic to estimate the advantage in the policy gradient term of the outer loss:
\begin{equation}
\label{eq:outer_advantage_bias}
     \hat{g}_\textrm{pg}'(\tau) = \left(r + \gamma' \textcolor{red}{V^{\pi,\gamma'}}(s')\right) \nabla_{\theta'} \log \pi_{\theta'}(a|s) \not\approx \left(r + \gamma' \textcolor{red}{V^{\pi,\gamma}}(s')\right) \nabla_{\theta'} \log \pi_{\theta'}(a|s)
\end{equation}
The right-hand side of equation~\ref{eq:outer_advantage_bias} is the biased estimate of the outer policy gradient used by current self-tuning methods.

\textbf{Removing the bias with an outer-critic head}: We propose to learn another critic that we call the \emph{outer-critic}, whose goal is to estimate the \emph{outer} value function, i.e. the value of the policy with (constant) discount $\gamma'$.
More precisely, the outer-critic $V^{\pi,\gamma'}_{\theta_{\bar{c}}}$, parameterized by $\theta_{\bar{c}}$, is trained to approximate the outer value function 
: 
$
    V^{\pi,\gamma'}(s) = \mathbb{E}_{\pi | s_t = s} \left[\sum_{k=0}^{\infty} {\gamma'}^k r_{t+k}\right] .
$
This way, one can use the outer-critic to remove the aforementioned bias in the estimation of the outer-loss: $\left(r + \gamma' V^{\pi,\gamma'}_{\theta_{\bar{c}}}(s')\right) \log \pi_{\theta'_p}(a|s)$.
To implement the outer-critic, we modify current critic architectures by adding a second head whose goal is to learn the outer value function $V^{\pi,\gamma'}$.
As a consequence, our way of removing the bias in the outer loss advantage estimation is computationally cheap and easy to implement on top of current self-tuning methods to improve meta-gradient estimation.
Further details are provided in  appendix~\ref{appendix:outer-critic}.

\section{Experiments}
We demonstrate the catastrophic failure of current meta-gradient approaches in a toy problem, then we scale to a deep RL experiment where we observe that our method improves performance over biased meta-gradients.
It is important to note that for these self-tuning experiments, we initialize $\gamma$ to a low value (0.95 and 0.8), in fact too low for a converged policy to be optimal (both experiments require optimizing over long horizons). This is to accentuate the effect of the meta-gradient bias.
We open-source the code to reproduce all the experiments, see \url{https://github.com/instadeepai/outer-value-function-meta-rl}.

\subsection{Discounting Chain}

\begin{figure}[h]
     \centering
    \begin{subfigure}[b]{0.32\textwidth}
         \centering
        \includegraphics[width=\textwidth]{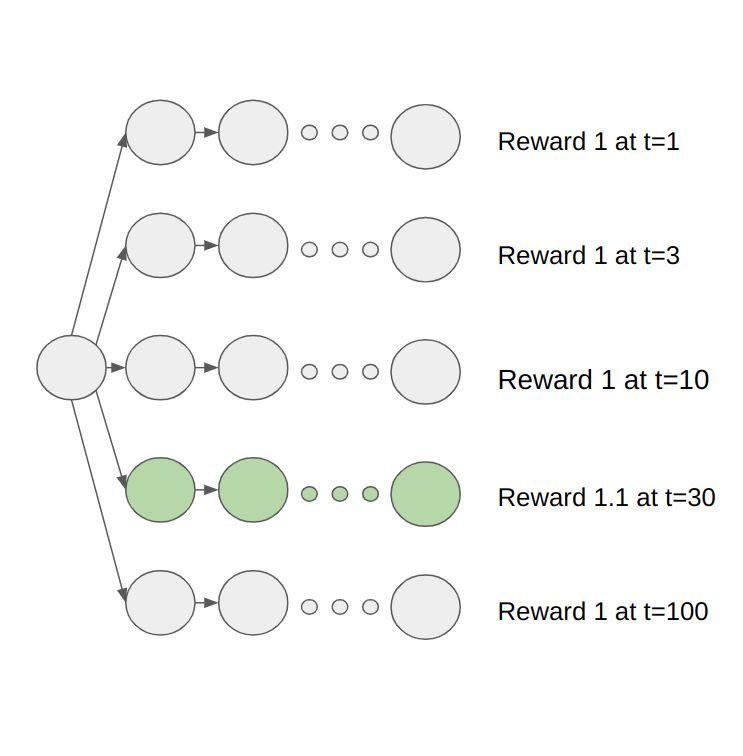}
    \end{subfigure}
    \hfill
     \begin{subfigure}[b]{0.32\textwidth}
         \centering
         \includegraphics[width=\textwidth]{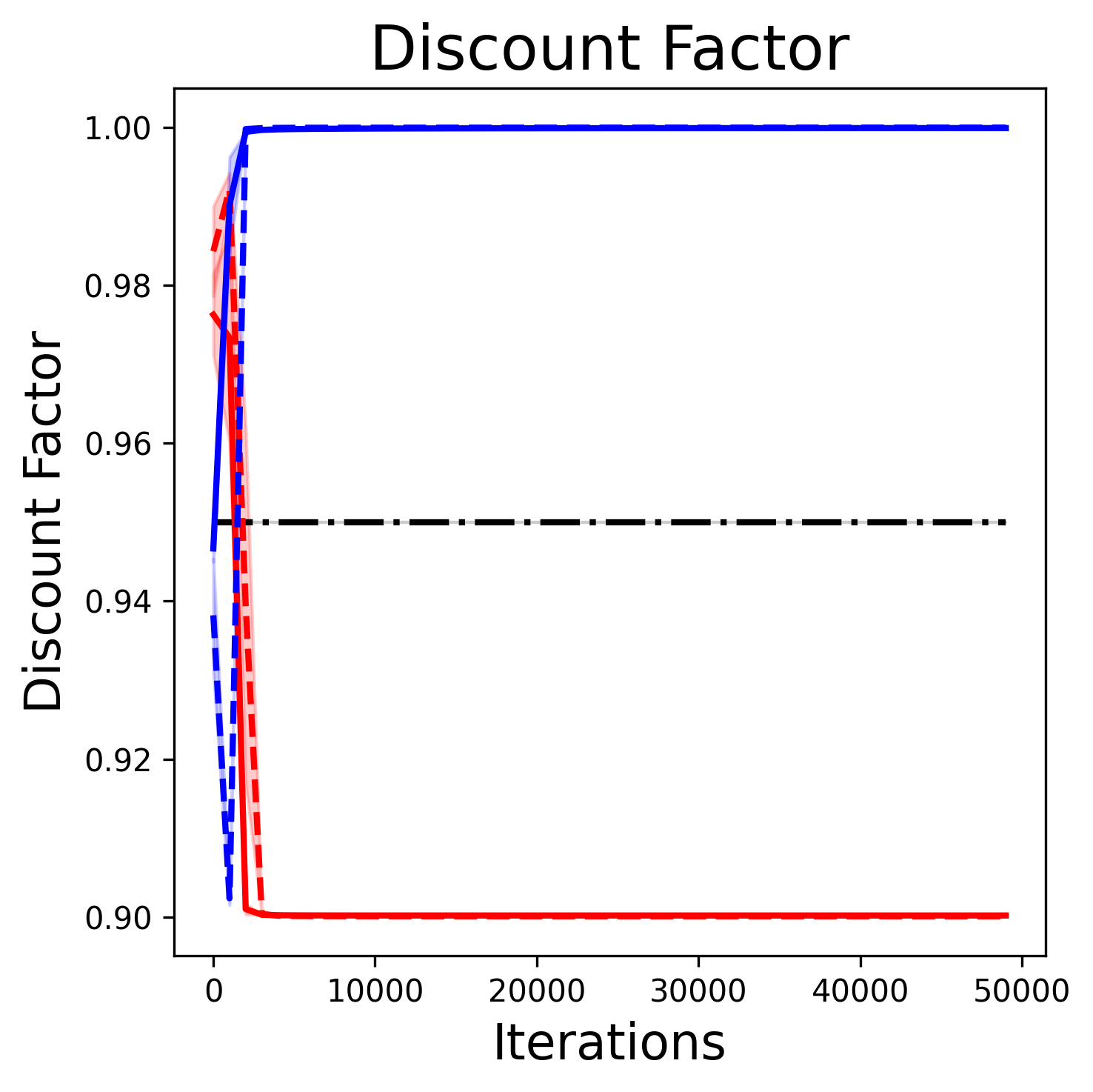}
     \end{subfigure}
     \hfill
     \begin{subfigure}[b]{0.32\textwidth}
         \centering
         \includegraphics[width=\textwidth]{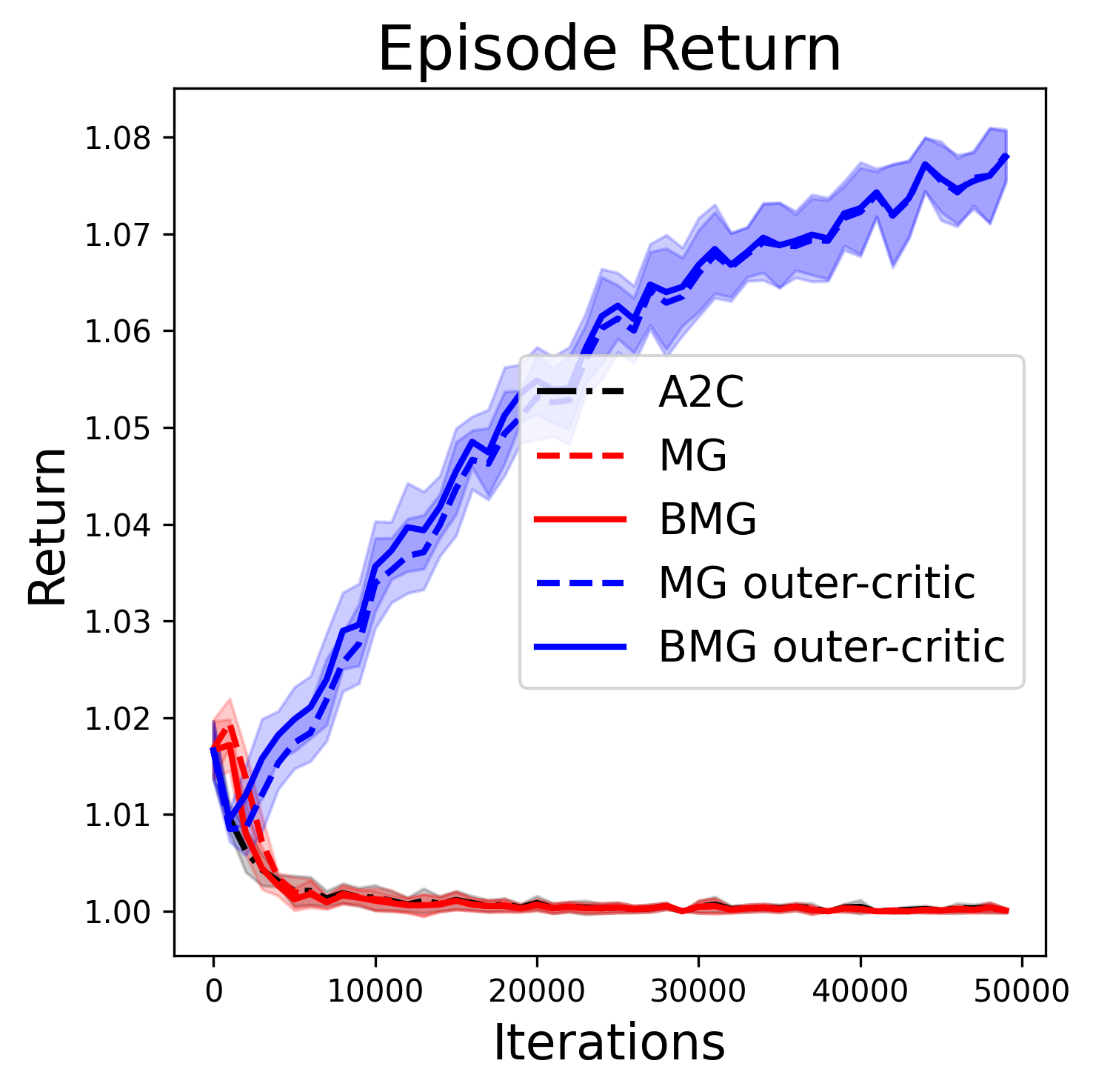}
     \end{subfigure}
    \caption{Discounting chain experiment. \textbf{(left)} A visualization of the environment. \textbf{(middle)} Discount during training. \textbf{(right)} Return during training (shares the legend with (middle)). Red curves show current self-tuning methods (MG and BMG) while blue curves are obtained by using an outer-critic to estimate the outer loss. Shaded areas correspond to one standard deviation across 10 random seeds.}
    \label{fig:dc_experiments}
    
\end{figure}

For our first experiment, we provide a simple illustration of the effects of the bias caused by using the inner critic in the outer loss. 
For this purpose, we experiment on the Gymnax~\cite{gymnax2022github} JAX implementation of the toy Discounting Chain environment originally proposed in bsuite\footnote{
We use the environment as an illustrative case rather than according to the specification of the bsuite setup.}
~\cite{Osband2020Behaviour}.
The Discounting Chain environment, shown in Figure \ref{fig:dc_experiments}, has a structure designed to highlight the effect of different discount horizons and is therefore highly sensitive to the discount factor.
Since we have access to the true value function for this environment, we use it in the policy gradient advantage when training the policy.
Having the true value function allows us to focus on the bias induced by using the wrong value of $\gamma$ in the outer loss.
We train the MG and BMG agents with and without the outer critic, with an initial $\gamma=0.95$.
We apply a modified sigmoid activation between $0.9$ and $1.0$ to ensure the discount remains bounded.
We provide further details on the experimental setup in Appendix \ref{appendix:discount-chain}. 
In figure~\ref{fig:dc_experiments}, the standard MG and BMG agents fail catastrophically, as they are unable to increase the discount factor and therefore converge to the myopic policy.
The reason is that if the inner-critic is used in the outer-loss advantage, then myopic policies are preferred (because $V^{\pi,\gamma} < V^{\pi,\gamma'}$), which then causes the meta-gradient to push $\gamma$ down instead of up.
On the other hand, the MG and BMG agents equipped with the outer critic are able to successfully self-tune the discount factor to converge to the optimal policy and solve the task. 

\subsection{Snake}

\begin{figure}[h]
     \centering
    \begin{subfigure}[b]{0.32\textwidth}
         \centering
    \includegraphics[width=\textwidth]{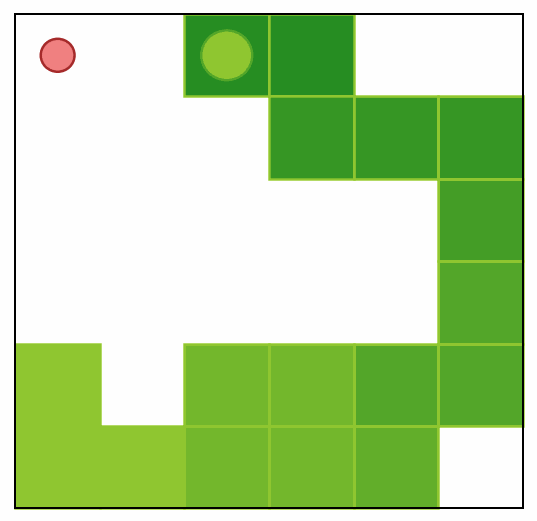}
    \end{subfigure}
    \hfill
     \begin{subfigure}[b]{0.32\textwidth}
         \centering
         \includegraphics[width=\textwidth]{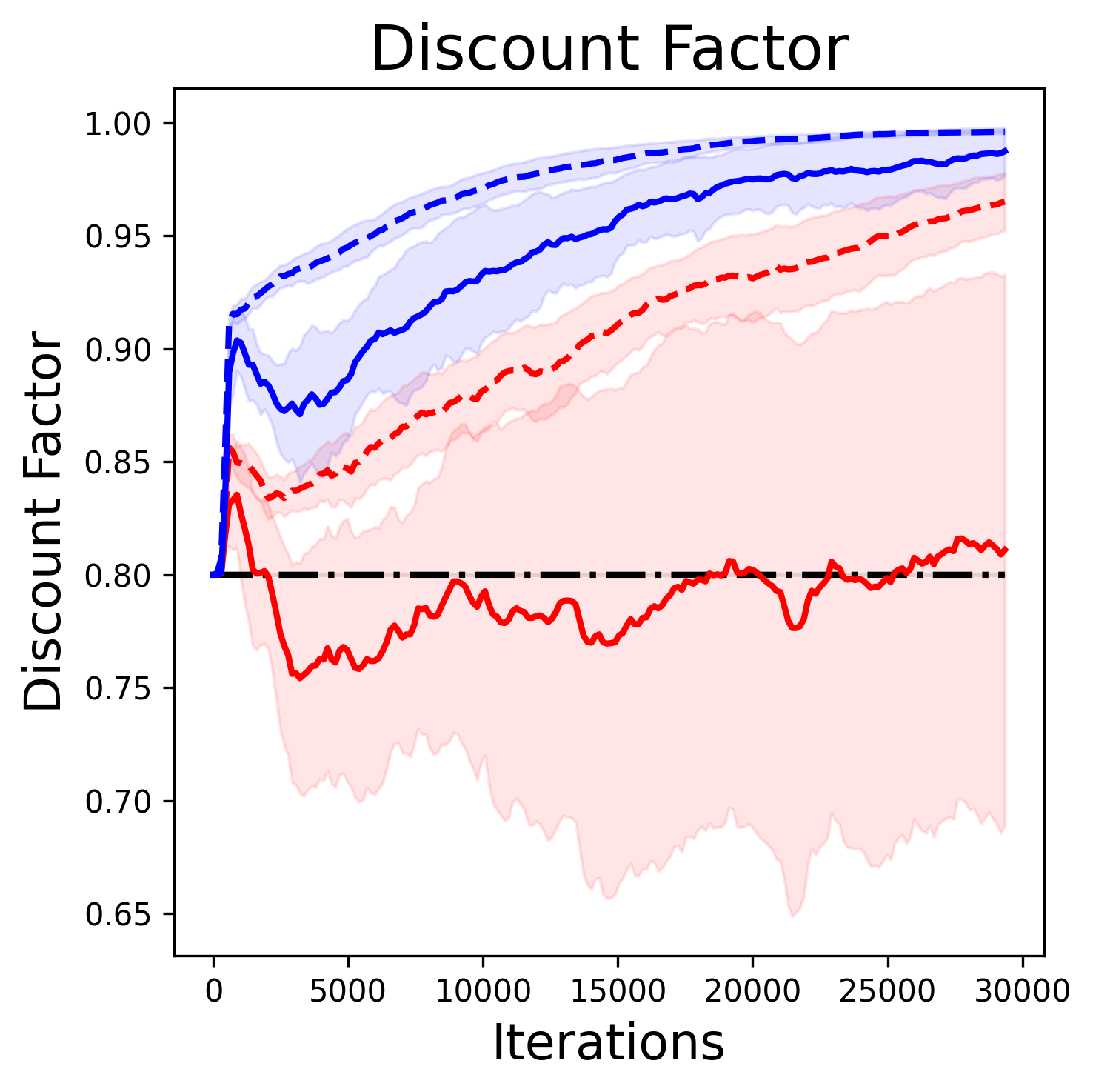}
     \end{subfigure}
     \hfill
     \begin{subfigure}[b]{0.32\textwidth}
         \centering
         \includegraphics[width=\textwidth]{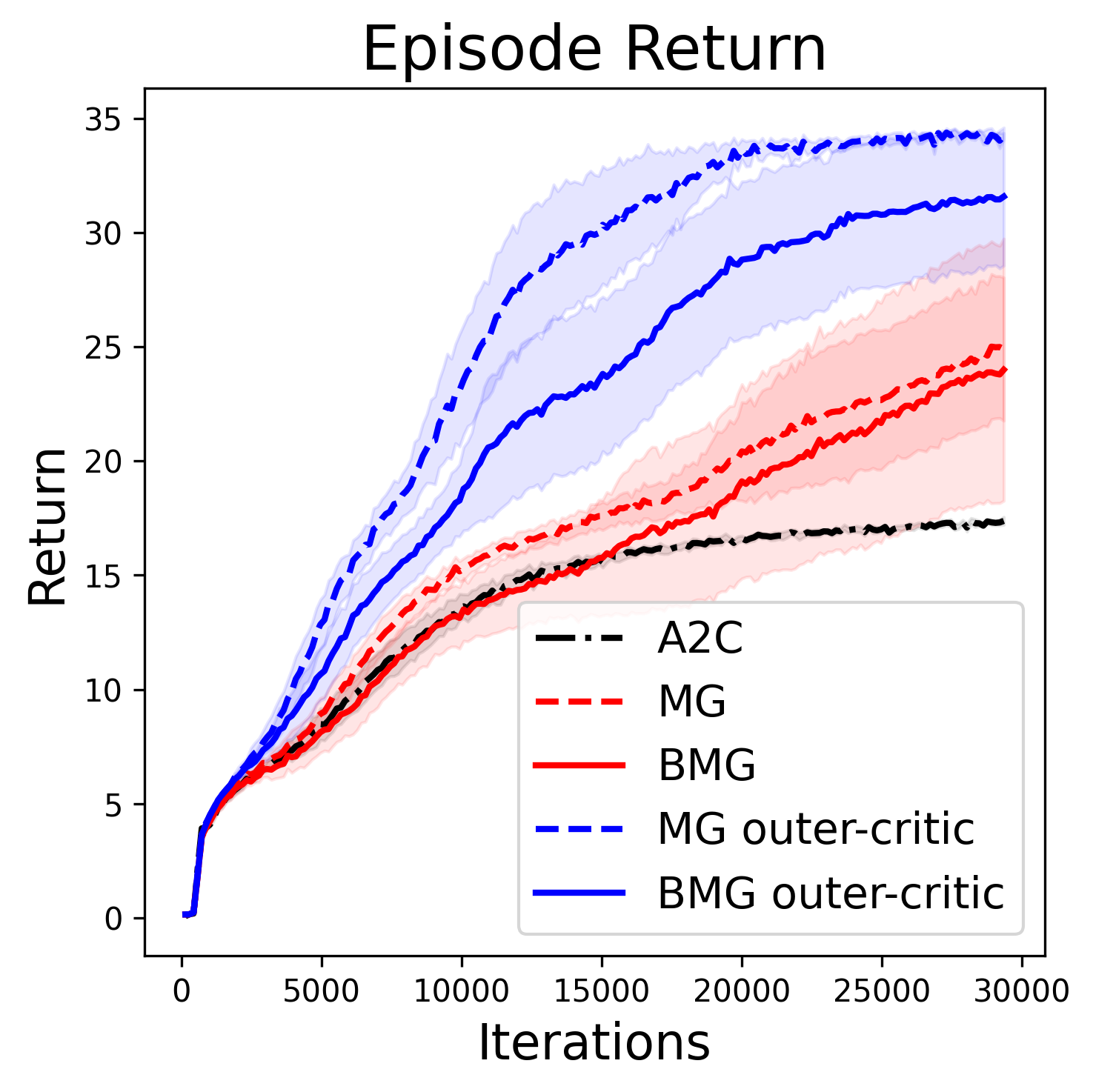}
     \end{subfigure}
    \caption{Snake experiment. (a) A visualization of the environment. (b) Discount during training. (c) Return during training. Shaded areas represent half a standard deviation across 10 seeds.}
    \label{fig:snake_experiments}
\end{figure}

Next, we scale our experiments to the deep RL setting using the Snake environment from Jumanji~\cite{jumanji2022github,bonnet2021one}.
This environment is sensitive to the discount factor, as the snake agent must plan ahead to maximize return without bumping into itself.
We compare an MG algorithm with and without the outer critic. 
Unlike in the Discounting Chain experiment, where we had access to the true value function, we now train a value function for the inner and outer critic. 
We give $\gamma$ a relatively poor initialization of $0.8$, as this makes the self-tuning more challenging.
We provide further details for the setup of the experiment in Appendix~\ref{appendix:snake}. 
In figure~\ref{fig:snake_experiments}, the agent with the second critic shows a stronger meta-gradient signal, which results in $\gamma$ being increased faster. 
Although both agents asymptotically converge to the maximum return, fixing the meta-gradient bias results in a significant improvement in training performance.

\section{Conclusion}

We have shown that current meta-gradient RL algorithms suffer from a meta-gradient bias induced by having the critic, trained using the meta-learned discount factor, estimate the advantage of the outer objective which necessitates a different discount factor.
In our experiments, we have demonstrated that this meta-gradient bias can lead to catastrophic failure of meta-gradient RL algorithms, which we fix with a simple, yet efficient solution.
We augment the critic network with an outer critic head whose goal is to estimate the value function discounted by the outer discount factor.
Furthermore, we have shown that our method results in performance improvements in more complex environments. 

Bootstrapped meta-gradient~\cite{flennerhag2022bootstrapped}, the state-of-the-art meta-gradient RL agent, has worse than human performance on Bowling, Solaris, and Skiing from the Atari ALE benchmark.
These environments are known to be challenging due to delayed reward/credit assignment~\cite{agent57,RUDDER}.
Thus, correcting the bias we highlight in this work may be important for improving meta-gradient RL approaches in these increasingly challenging environments, which we hope to study in future work.

\section*{Acknowledgements}
Research supported with Cloud TPUs from Google's TPU Research Cloud (TRC). We would like to thank Nathan Grinsztajn for their helpful discussions.

\bibliographystyle{abbrv}
\bibliography{references} 

\newpage
\appendix

\section{Background}
\label{appendix:background}

The goal of an RL agent is to maximize the $\gamma$-discounted state value function: 
$V^{\pi,\gamma}(s) = \mathbb{E}_{\pi | s_t = s} \left[\sum_{k=0}^{\infty} \gamma^k r_{t+k}\right]$.

The action value function is analogously defined below.
\begin{align*} 
    Q^{\pi,\gamma}(s, a)
    &= \mathbb{E}_{\pi | s_t = s, a_t = a} \left[\sum_{k=0}^{\infty} \gamma^k r_{t+k}\right] \\
    &= \mathbb{E}_{\pi | s_t = s, a_t = a} \left[r_t + \gamma \sum_{k=0}^{\infty} \gamma^k r_{t+1+k}\right] \\
    &= \mathbb{E}_{\pi | s_t = s, a_t = a} \left[r_t + \gamma V^{\pi,\gamma}(s_{t+1})\right] \\
\end{align*}

The policy gradient theorem~\cite{Williams1992,gae_paper} in equation~\ref{eq:policy_gradient_theorem} provides a way to derive the gradient of the expected discounted return using the value function of the policy.
\begin{equation}
\begin{aligned} 
\label{eq:policy_gradient_theorem}
    \nabla_\theta \mathbb{E}_{\pi_\theta}\left[\sum_t \gamma^t r_t\right]
    &= \mathbb{E}_{\pi_\theta}\left[\sum_t \gamma^t r_t \nabla_\theta \log \pi_\theta(a_t|s_t) \right] \\
    &= \mathbb{E}_{\pi_\theta}\left[\sum_t Q^{\pi,\gamma}(s_t, a_t)  \nabla_\theta \log \pi_\theta(a_t|s_t) \right] \\
    &= \mathbb{E}_{\pi_\theta}\left[\sum_t \left(r_t + \gamma V^{\pi,\gamma}(s_{t+1})\right) \nabla_\theta \log \pi_\theta(a_t|s_t) \right] \\
\end{aligned}
\end{equation}

Actor-critic algorithms~\cite{konda1999,mnih2016} learn a parameterized value function (the critic) and a parameterized policy which is updated in a direction informed by the critic. 
More precisely, the critic learns the value function $V^{\pi,\gamma}_{\theta_c}$ with critic parameters $\theta_c$, and policy $\pi_{\theta_p}$ with policy parameters $\theta_p$.
Estimated action value function $\hat{Q}^{\pi,\gamma}(s, a) = r + \gamma V^{\pi,\gamma}_{\theta_c}(s')$ where $r$ and $s'$ are respectively the reward and the next state obtained from taking action $a$ in state $s$.
The goal of an RL agent is to learn the policy $\pi$ that maximizes the $\gamma$-discounted expected return:  $\mathbb{E}_{\pi | s_t = s} \left[\sum_{k=0}^{\infty} \gamma^k r_{t+k}\right]$, where $r_t$ and $s_t$ are the reward and state at timestep $t$.

An actor-critic update (A2C)~\cite{mnih2016} is composed of 3 terms:
\begin{itemize}
    \item Policy gradient: $\left(r + \gamma V^{\pi,\gamma}_{\theta_c}(s')\right) \nabla_{\theta_p} \log \pi_{\theta_p}(a|s)$
    \item Critic gradient: $\left(V^{\pi,\gamma}_{\theta_c}(s) - \left(r + \gamma V^{\pi,\gamma}_{\theta_c}(s')\right)\right) \nabla_{\theta_c} V^{\pi,\gamma}_{\theta_c}(s)$
    \item Entropy regularization: $\nabla_{\theta_p} H(\pi_{\theta_p}(.|s))$
\end{itemize}

Therefore, the full actor-critic update is: $\theta \leftarrow \theta + \alpha f(\theta, \eta, \tau)$ where $\theta = \{\theta_p, \theta_c\}$ and $\eta = \{\gamma\}$ or $\eta = \{\gamma, \lambda\}$ if TD($\lambda$) is used for advantage estimation, and

\begin{multline*}
    f(\theta, \eta, \tau) = c_\textrm{PG}\left(r + \gamma V^{\pi,\gamma}_{\theta_c}(s')\right) \nabla_{\theta_p} \log \pi_{\theta_p}(a|s) \\
    + c_\textrm{TD}\left(V^{\pi,\gamma}_{\theta_c}(s) - \left(r + \gamma V^{\pi,\gamma}_{\theta_c}(s')\right)\right) \nabla_{\theta_c} V^{\pi,\gamma}_{\theta_c}(s) + c_\textrm{EN}\nabla_{\theta_p} H(\pi_{\theta_p}(.|s))
\end{multline*}

\section{Outer Critic}
\label{appendix:outer-critic}

\begin{figure}[h]
    \begin{center}\includegraphics[width=0.7\textwidth]{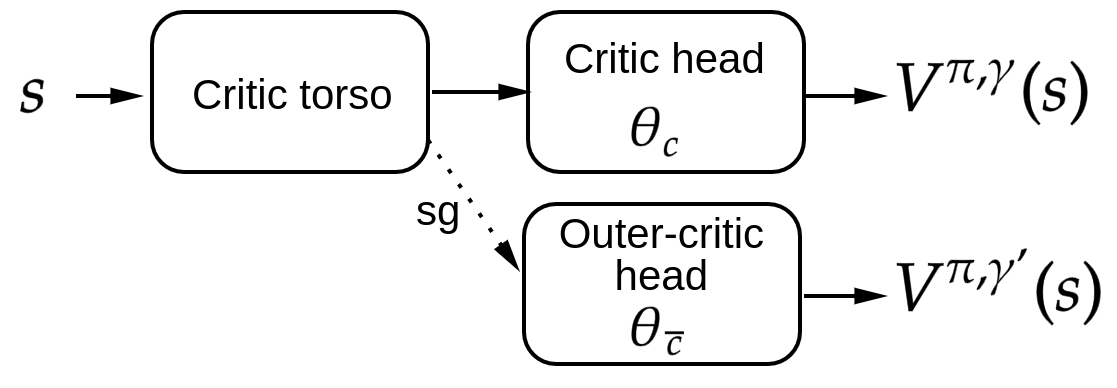}
    \end{center}
    \caption{Augmented critic architecture to learn two value functions: the inner and outer values. "sg" means that the \texttt{stop\_gradient} operation is used to prevent gradients from flowing from the outer-critic head to the common torso.}
    \label{fig:outer-critic-head}
\end{figure}

To learn the outer value function, we augment the critic network with a second head that outputs the value with discount $\gamma'$. To avoid conflicting gradients between the two value functions being learned, we use a stop-gradient between the common torso and the outer-critic head (see figure~\ref{fig:outer-critic-head}). This way, the critic learning dynamic is not changed by adding the outer-critic.

To learn the outer value function, we proceed akin to the standard value function with a TD loss $l_\textrm{TD}^\textrm{outer}$ with $\gamma'$ instead of $\gamma$.
\begin{equation*}
    l_\textrm{TD}^\textrm{outer} = \left(V_{\theta_{\bar c}}^{\pi, \gamma'}(s) - \left(r + \gamma' \overline{V_{\theta_{\bar c}}^{\pi, \gamma'}(s')}\right)\right)^2
\end{equation*}

\section{Experiments}

\subsection{Discounting Chain Experiment Details}
\label{appendix:discount-chain}

We initialize $\gamma=0.95$ such that the value function will initially favor the myopic policy. 
For simplicity, we use only a policy gradient loss for both the inner and outer loss functions.
Furthermore, for BMG we do a single gradient step with the outer loss to obtain the target, which we use for policy matching. 
We optimize the parameters of the actor for MG and BMG with SGD and use Adam~\cite{kingma2015} to optimize the discount factor $\gamma$.

For all agents, we use a policy network composed of a single linear layer.
We generate a new batch of data online for both the inner and outer loss functions.
Using an adaptive optimizer (e.g. Adam or RMSProp) to optimize the parameters would change the meta-gradient for BMG since there would be some inertia from the inner update biasing the direction of the target update.
As we would like our inner update step to be in the direction of the unbiased meta-gradient, we use SGD instead of Adam.
More details on the hyper-parameters used in this experiment can be found in table~\ref{tab:experiments_hyperparameters}.

To solve the environment, the self-tuning agent has to increase the discount factor to a sufficient level so that it can receive a signal from the long-term effect of the first action.
However, if the meta-objective is sufficiently biased, the meta-gradient will not encourage the agent to increase the discount. 

\subsection{Snake Experiment Details}
\label{appendix:snake}

We use the environment "Snake-6x6-v0" from Jumanji~\cite{jumanji2022github} by calling Jumanji's registry: \texttt{jumanji.make("Snake-6x6-v0")}.
In this environment, a \emph{snake} agent navigates a grid in order to eat apples, while avoiding bumping into its own body which causes it to die.
Because good policies require the agent to zigzag and reason about long horizons, the discount factor plays a very important role in the learning dynamics and is well suited to studying any meta-gradient biases.
The hyper-parameters used in the experiments are provided in table~\ref{tab:experiments_hyperparameters}.

\begin{table}[ht]
\centering
\begin{tabular}{c|c|c}
    \hline
    \textbf{Parameters} & \textbf{Discounting Chain} & \textbf{Snake} \\ [0.5ex]
    \hline
    architecture & Linear & Conv + MLP \\
    $\gamma^{\textrm{start}}$ & 0.95 & 0.8 \\
    $\lambda$ & 0.0 & 0.95 \\
    $c_\textrm{PG}$ & 1.0 & 1.0 \\
    $c_\textrm{TD}$ & 0.0 & 0.5 \\
    $c_\textrm{EN}$ & 0.005 & 0.01 \\
    learning rate & 0.5 & 5e-4 \\
    inner optimiser & SGD & RMSProp \\
    gradient clipping norm & None & None \\
    batch size & 128 & 512 \\ [0.5ex]
    sequence length & 100 & 5 \\ [0.5ex]
    \hline
    $\gamma'$ & 1.0 & 1.0 \\
    $\lambda'$ & 0.0 & 1.0 \\
    $c_\textrm{PG}'$ & 1.0 & 1.0 \\
    $c_\textrm{TD}'$ & 0.0 & 0.0 \\
    $c_\textrm{EN}'$ & 0.005 & 0.0 \\
    MG meta-learning rate & 0.1 & 3e-3 \\
    BMG meta-learning rate & 0.1 & 6e-3 \\
    meta optimiser & Adam & Adam \\
    meta-gradient clipping norm & None & 0.1 \\
    \hline
\end{tabular}
\caption{Hyper-parameters used in the experiments.}
\label{tab:experiments_hyperparameters}
\end{table}

\section{Outer Loss Advantage}

\subsection{Quantify the Meta-Gradient Bias}

\begin{figure}[h]
     \centering
    \begin{subfigure}[b]{0.45\textwidth}
         \centering
    \includegraphics[width=\textwidth]{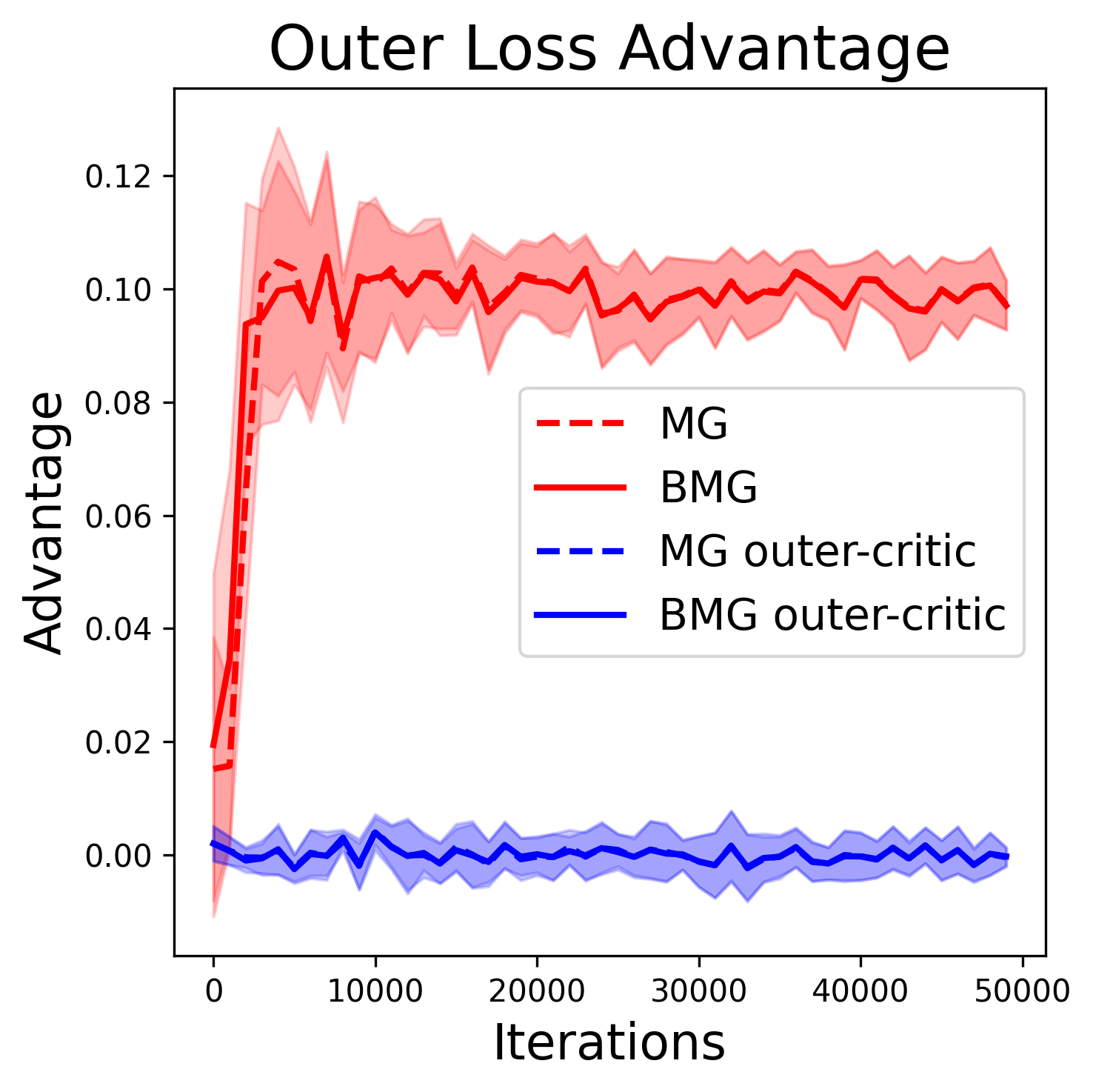}
         \caption{Discounting Chain}
    \end{subfigure}
    \hfill
     \begin{subfigure}[b]{0.45\textwidth}
         \centering
         \includegraphics[width=\textwidth]{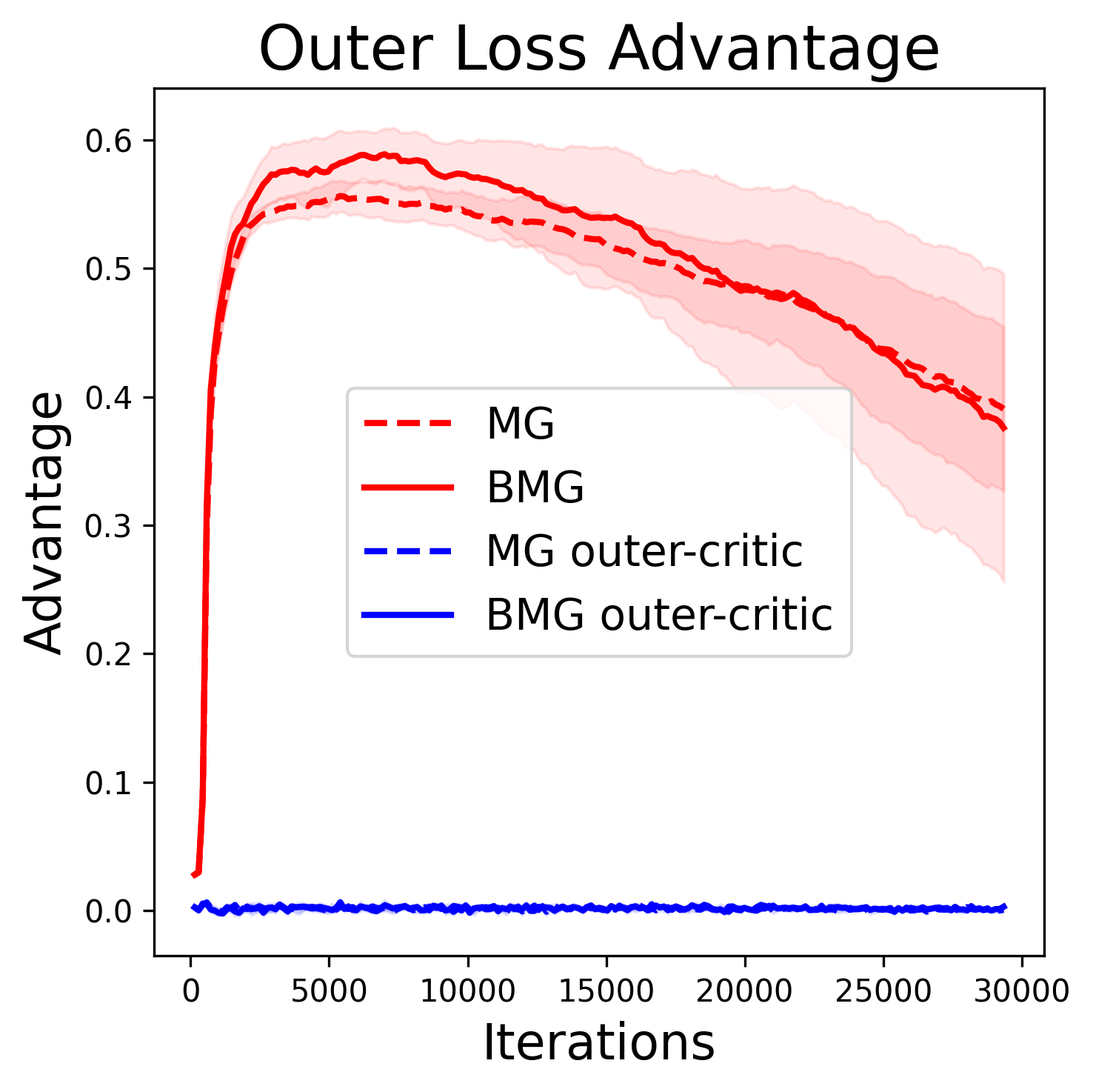}
         \caption{Snake}
     \end{subfigure}
    \caption{Advantage estimation in the outer loss using the standard critic versus the outer-critic. We see that the bias caused by using the standard critic in the outer loss results in the advantage being biased such that it is no longer centered on $0$ as it should be. Using the outer critic fixes this bias as the advantage is now centered around $0$.}
    \label{fig:outer_loss_advantage}
\end{figure}

One way to quantify the meta-gradient bias we highlight in this paper is to observe the advantage in the outer loss. By definition, the advantage is centered, i.e. its expectation over actions is 0. However, we can see in figure~\ref{fig:outer_loss_advantage} that current self-tuning methods (MG, BMG) suffer from a bias in their advantage estimation. Using the outer value function in the outer loss fixes the bias and the advantage remains centered.

\subsection{Normalizing Advantages}

We show that the advantages in the outer loss are not centered, which leads to a meta-gradient bias.
One could argue that normalizing the advantages across a batch would fix this issue.
We show in this section that this is not the case.

\begin{figure}[h]
     \centering
    \begin{subfigure}[b]{0.32\textwidth}
         \centering
    \includegraphics[width=\textwidth]{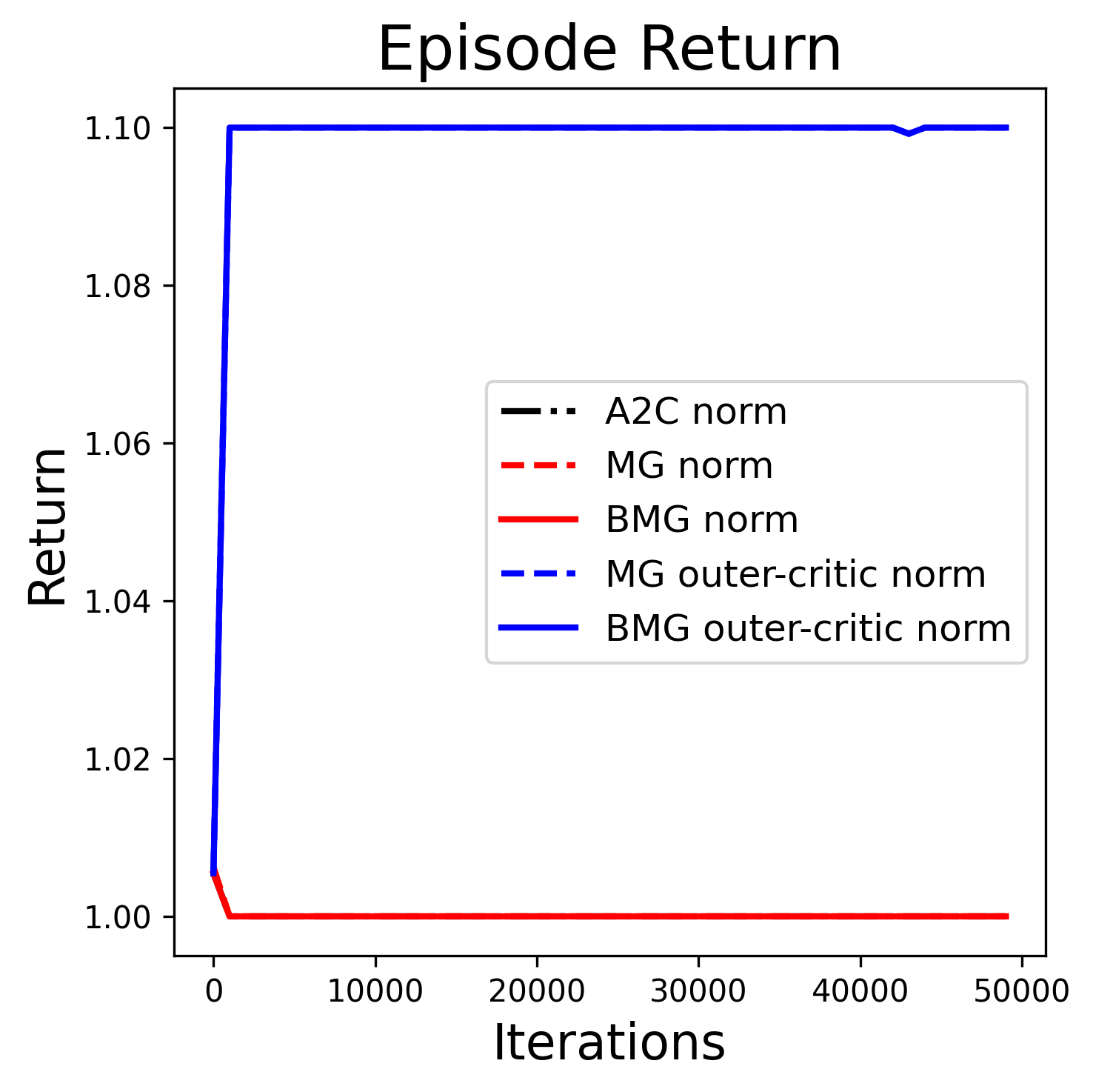}
    \end{subfigure}
    \hfill
     \begin{subfigure}[b]{0.32\textwidth}
         \centering
         \includegraphics[width=\textwidth]{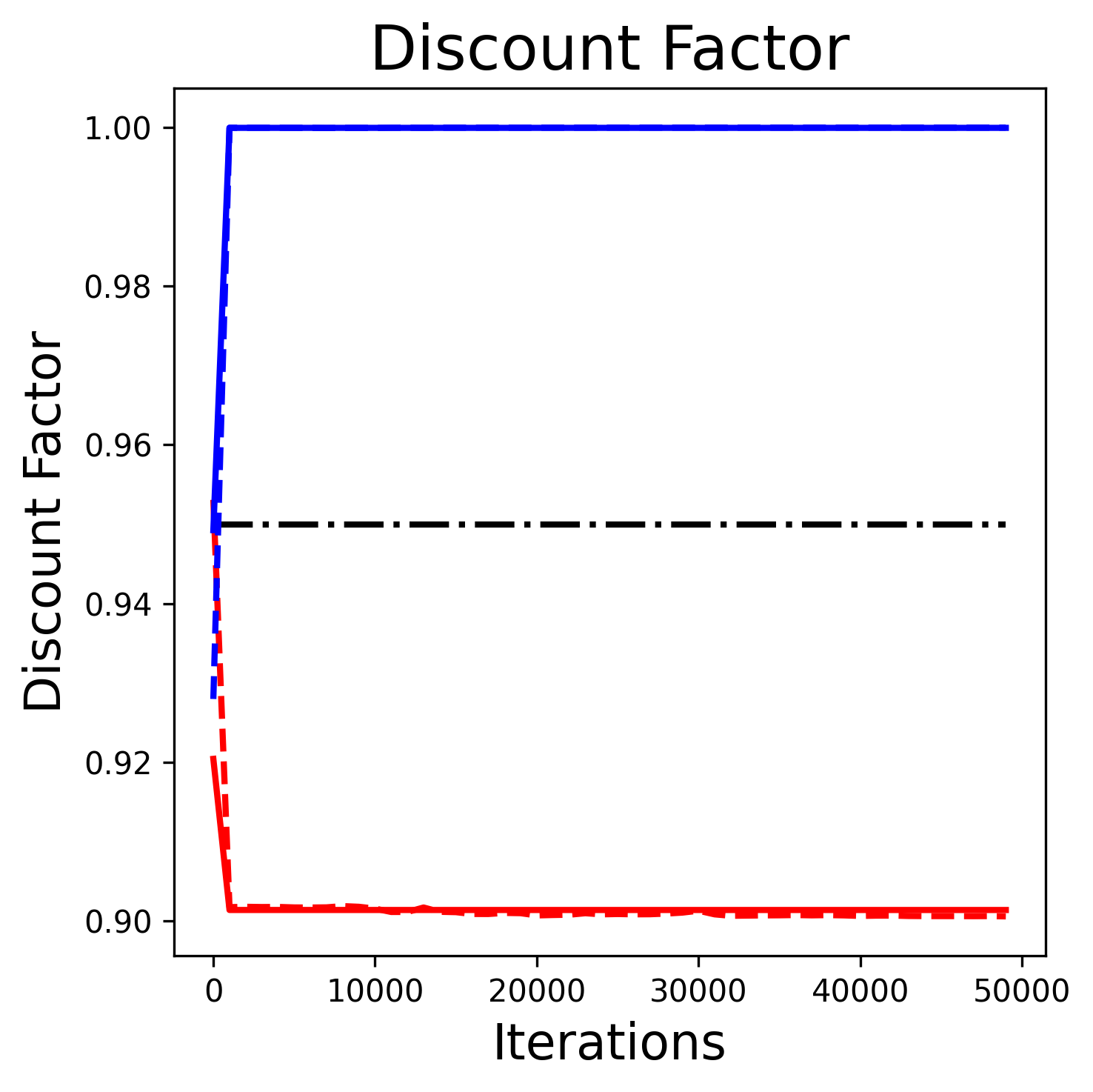}
     \end{subfigure}
     \hfill
     \begin{subfigure}[b]{0.32\textwidth}
         \centering
         \includegraphics[width=\textwidth]{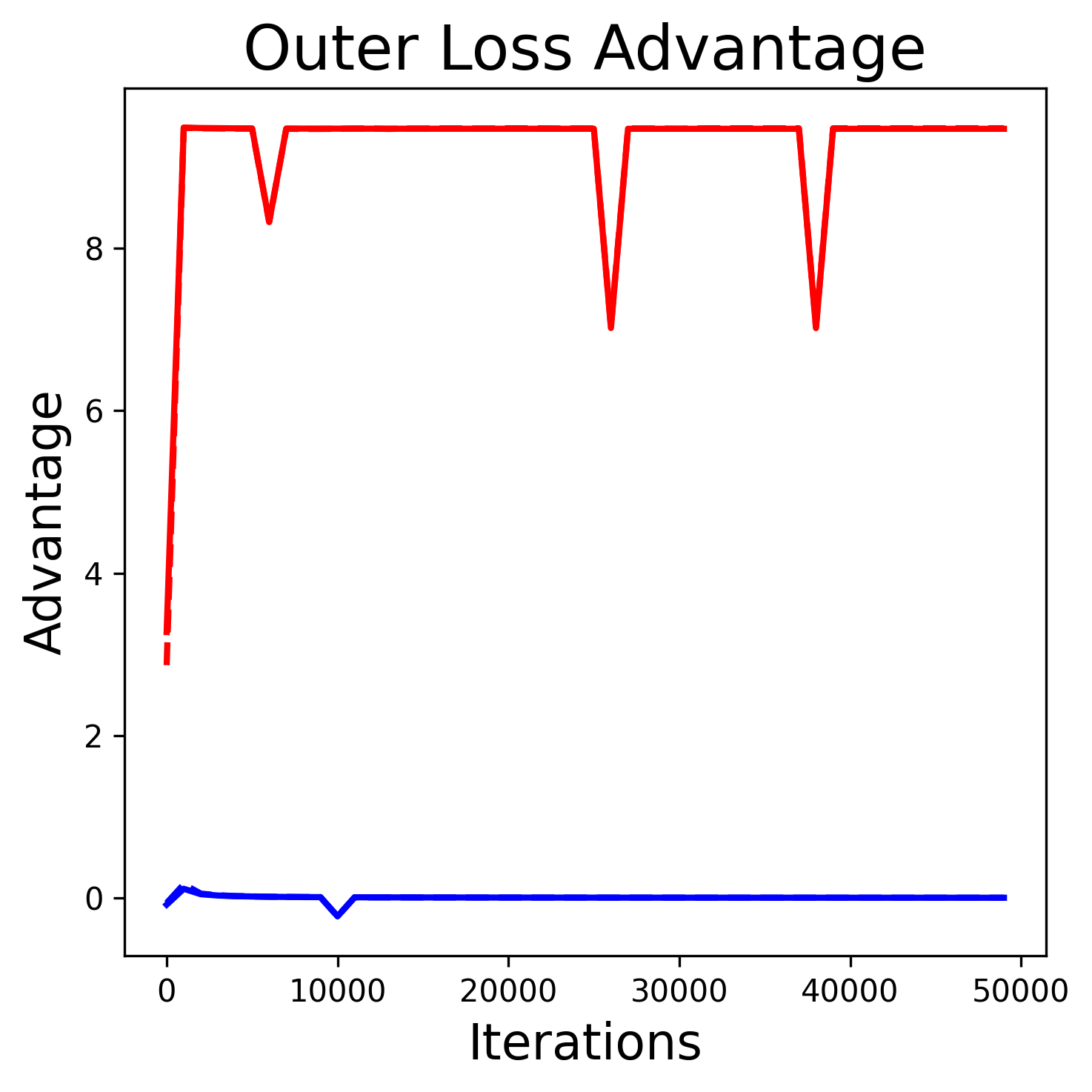}
     \end{subfigure} \\
     \centering
    \begin{subfigure}[b]{0.32\textwidth}
         \centering
    \includegraphics[width=\textwidth]{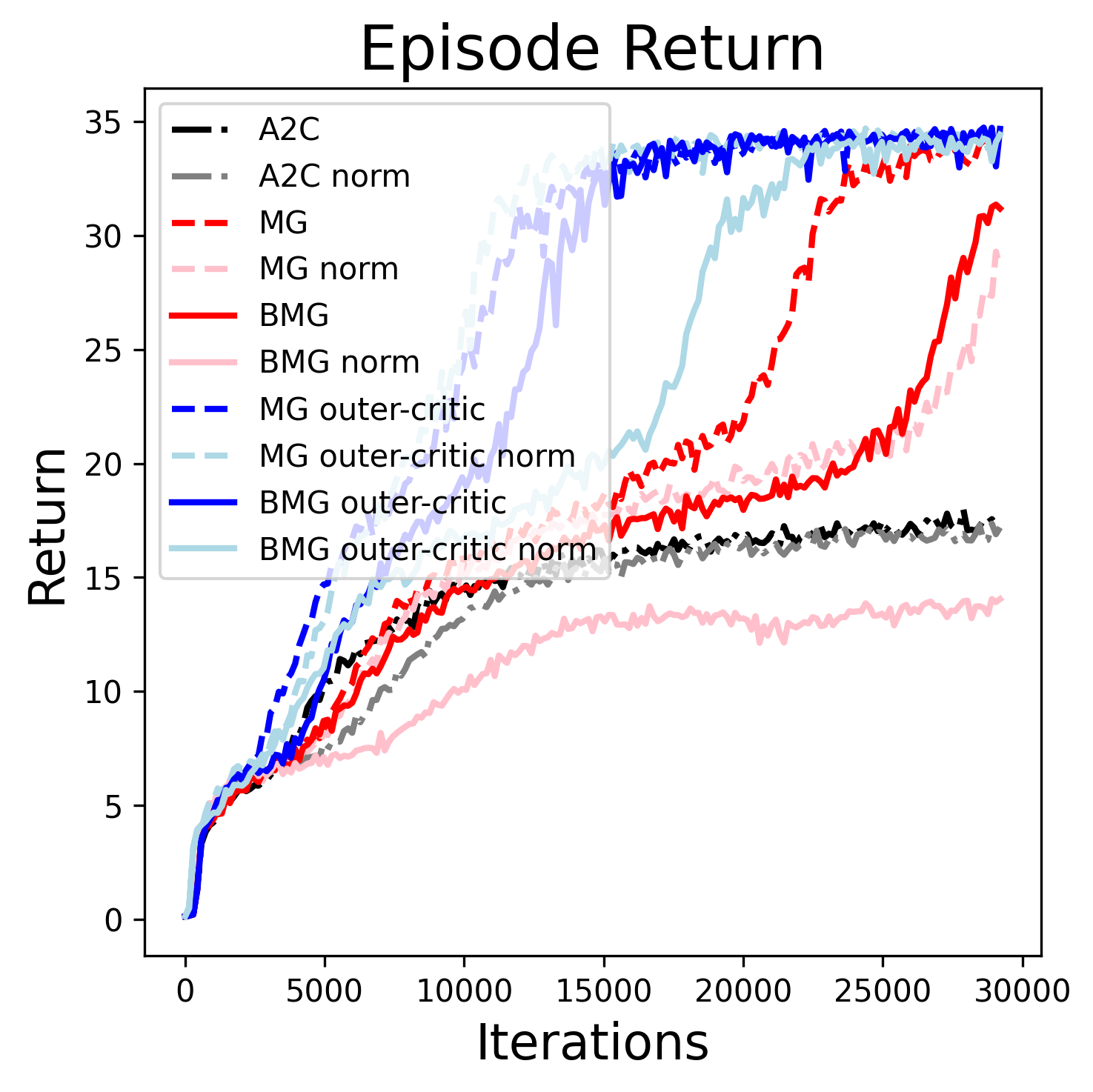}
    \end{subfigure}
    \hfill
     \begin{subfigure}[b]{0.32\textwidth}
         \centering
         \includegraphics[width=\textwidth]{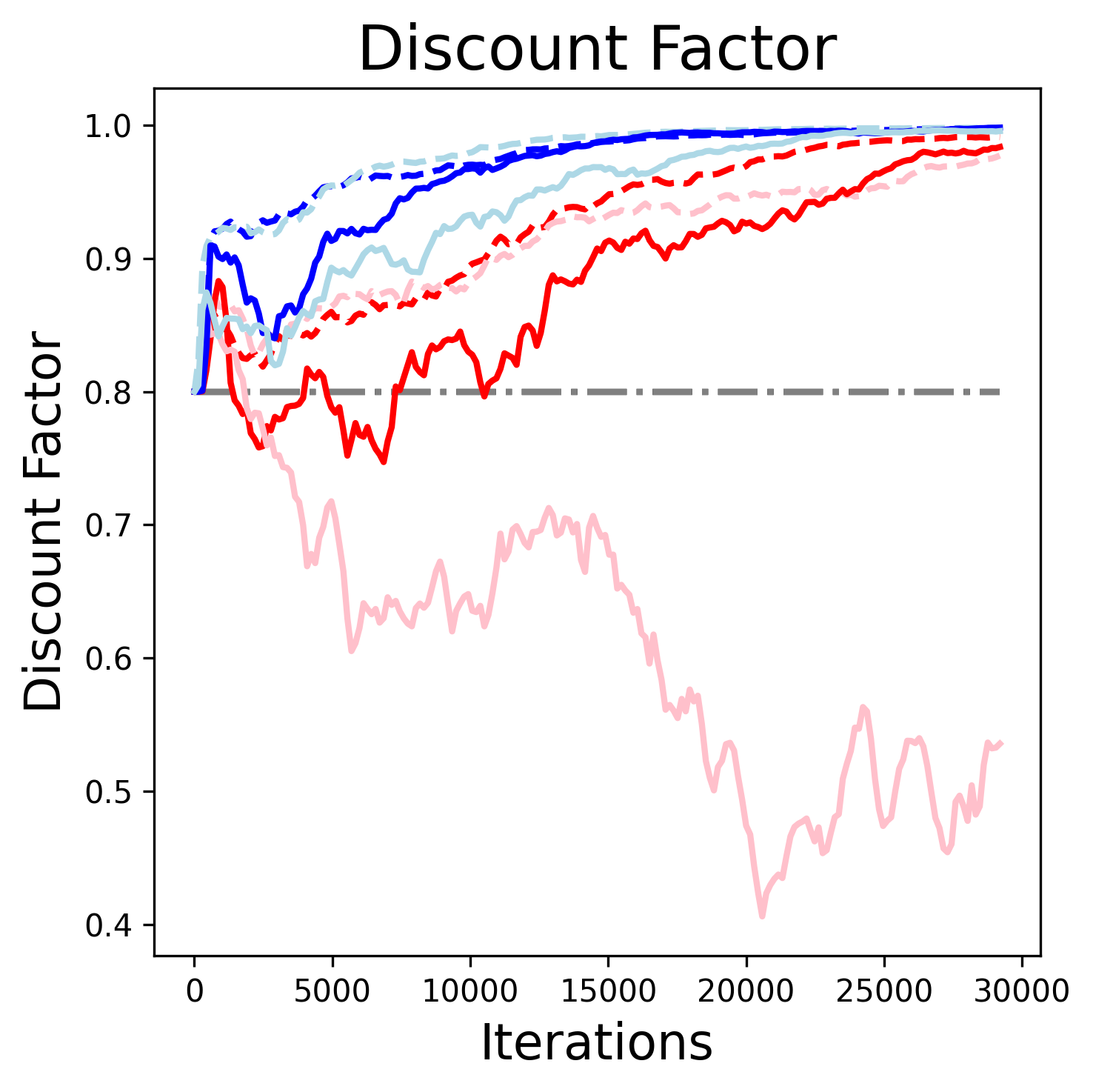}
     \end{subfigure}
     \hfill
     \begin{subfigure}[b]{0.32\textwidth}
         \centering
         \includegraphics[width=\textwidth]{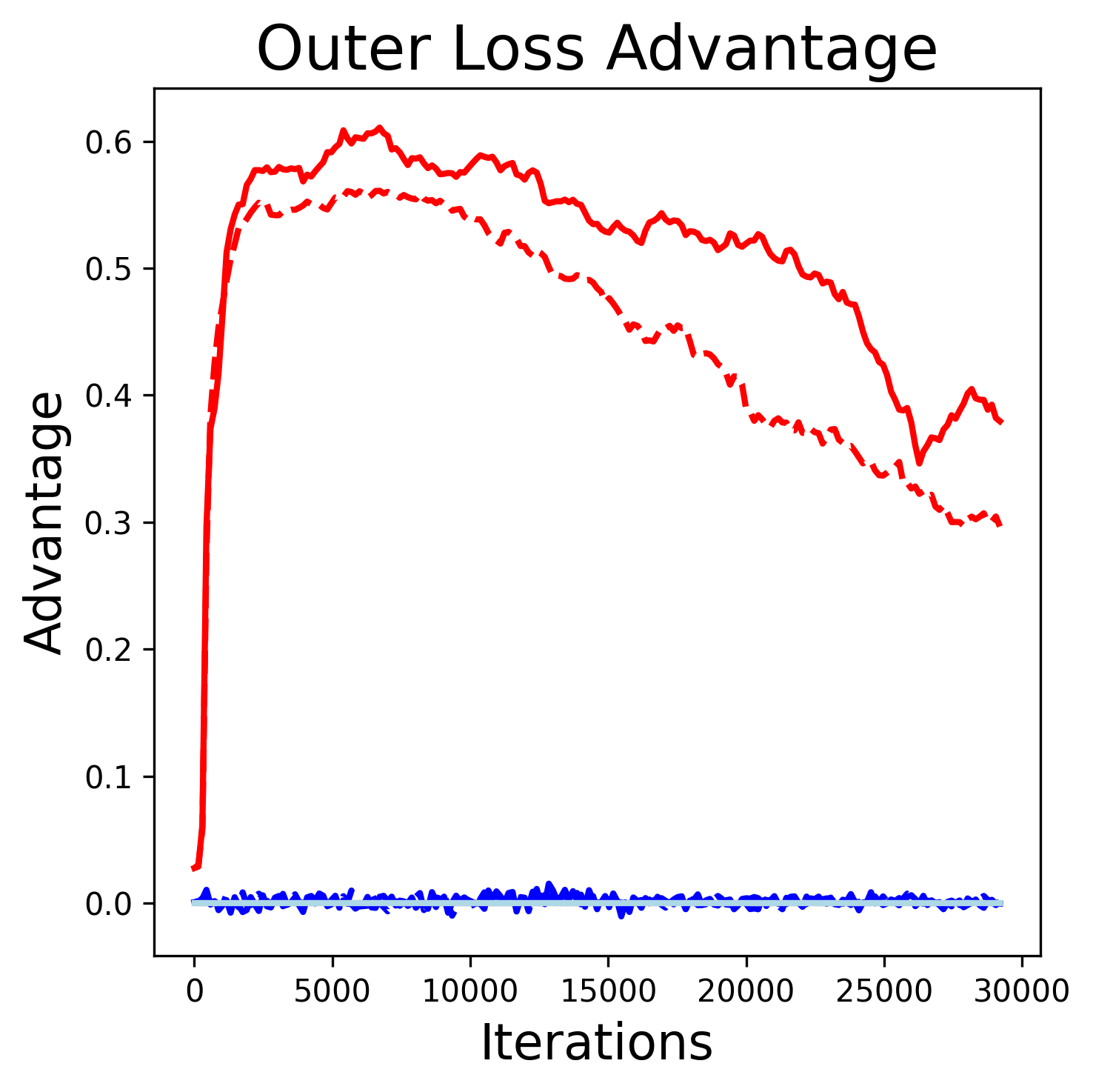}
     \end{subfigure}
    \caption{Experiments on both environments (Discounting Chain and Snake) with advantage normalization on 1 seed. (First row): Discounting Chain with all curves being with advantage normalization. (Second row): Snake with both advantage normalization or not. (Left): mean episode return during training. (Middle): meta-parameter curves. (Right): outer loss advantage. We observe that normalizing the advantage is misleading because it does not fix the meta-gradient though it shows a centered outer loss advantage.}
    \label{fig:advantage_normalization_exp}
\end{figure}

Indeed, in a batch, all the samples' advantages are biased. Removing this offset on the batch level does not fix the direction of the meta-gradient. This leads to the normalized advantages being centered but the meta-gradient still keeps its bias as shown in figure~\ref{fig:advantage_normalization_exp}.

\end{document}